\documentclass[sigconf]{acmart}
\usepackage[english]{babel}
\usepackage[utf8]{inputenc}
\usepackage{algorithm}
\usepackage[noend]{algpseudocode}
\AtBeginDocument{%
  \providecommand\BibTeX{{%
    \normalfont B\kern-0.5em{\scshape i\kern-0.25em b}\kern-0.8em\TeX}}}
\setcopyright{acmcopyright}
\copyrightyear{2018}
\acmYear{2018}

\acmConference[TRAITS 2021]{2021 ACM/IEEE International Conference on Human-Robot Interaction : AI, Trust and ethicS Workshop}{March 08--12, 2021}{Virtual}

\begin{document}

\title{Risk-Averse Biased Human Policies in Assistive Multi-Armed Bandit Settings}



\author{Michael Koller}
\email{koller@acin.tuwien.ac.at}
\affiliation{%
  \institution{ACIN, 
  TU Wien}
  \streetaddress{Gußhausstraße 27-29}
  \country{Austria}
  \postcode{1040}
}

\author{Timothy Patten}
\email{patten@acin.tuwien.ac.at}
\affiliation{%
  \institution{ACIN, 
  TU Wien}
  \streetaddress{Gußhausstraße 27-29}
  \country{Austria}
  \postcode{1040}
}

\author{Markus Vincze}
\email{vincze@acin.tuwien.ac.at}
\affiliation{%
  \institution{ACIN, 
  TU Wien}
  \streetaddress{Gußhausstraße 27-29}
  \country{Austria}
  \postcode{1040}
}
\begin{abstract}
Assistive multi-armed bandit problems can be used to model team situations between a human and an autonomous system like a domestic service robot. To account for human biases such as the risk-aversion described in the Cumulative Prospect Theory, the setting is expanded to using observable rewards. When robots leverage knowledge about the risk-averse human model they eliminate the bias and make more rational choices. We present an algorithm that increases the utility value of such human-robot teams. A brief evaluation indicates that arbitrary reward functions can be handled.
\end{abstract}
\ccsdesc[500]{Mathematics of computing~Probabilistic inference problems}
\ccsdesc[500]{Computing methodologies~Cognitive robotics}
\keywords{Human-Robot Interaction, Theory of Mind, Multi-Armed Bandit}
\maketitle
\section{Introduction}
Humans frequently find themselves playing multi-armed bandit (MAB) games. In such settings, an actor repeatedly chooses an action (or pulls an arm)
without complete knowledge about the associated reward distribution of each action. During an episode, there is a trade-off between choosing what previously yielded the best results (exploitation) and choosing other actions to improve the estimate of the mean reward (exploration). The goal for the autonomous system is to improve the return for a human by estimating the true expected return of each action by monitoring the human's behavior as implicit feedback \cite{sakagami1997learning, li2010contextual}, e.g., when 
choosing music or repeatedly commanding a robot. 
We might interact with domestic robots in situations, where the human partner is not quite sure what the robot understands as it learns about the different outcomes.
%
Examples include under-specified commands such as ``Set the table.'' or ``Fetch me some food for lunch.''. 

This paper focuses on MAB scenarios that model these decision situations.
%
When humans make choices, however, it is known that systematic biases occur, which lead to sub-optimal behavior. One such bias, which occurs in risky uncertain situations, is modeled by the \textit{Cumulative Prospect Theory} (CPT) \cite{kahneman1979prospect, tversky1992advances}. Humans tend to prefer small gains with small variance to large gains with large variance, but are willing to take more risk to avoid big losses. This bias was applied to a human-robot interaction (HRI)  \cite{kwon2020humans}, where an episode consisted of a single choice, but was not explored in repeated games. 
On the other hand, human-robot teams in MAB settings have been studied in \cite{chan2019assistive, milli2017should},
but their human models were either (noisy) rational, greedy or inconsistent. The solutions would only learn to replicate a bias as proposed by CPT.

The risk-averse biased human MAB policy used in this work is also distinct from MAB policies that explicitly minimize a risk factor \cite{vakili2015mean, galichet2013exploration, maillard2013robust, cassel2018general}. The biased policy's only goal is to maximize return, but has a biased perception of the estimated means during play, which leads to suboptimal behavior. 
In this paper we (1) define the assistive MAB with observable reward classes, (2) explore the behavior of a risk-averse biased MAB policy, and (3) present an assistive algorithm for a human-robot team that eliminates the bias of the human and improves the overall return for the team. 
\section{Multi-Armed Bandit Formulation}
The \textbf{MAB} is a simplified reinforcement learning problem, where an actor repeatedly picks one of $N$ available actions. In this work, each action is associated with a separate distribution over $M$ different rewards, from which rewards are sampled. The agent thereby weighs exploration 
and exploitation to finally settle on choosing the best arm. 
The formalism is presented for discrete random reward variable distributions. 
Continuous distributions are handled by dividing the continuous reward range into discrete classes.

In the \textbf{Inverse Multi-Armed Bandit with Observable Reward Classes} problem, a passive agent estimates the mean reward of the different actions by observing the MAB choices of another agent. The observer is only allowed to observe the chosen action and the reward class, i.e., no numeric utility, but only a class label. 
The \textbf{Assistive Multi-Armed Bandit with Observable Reward Classes} is a version of the inverse MAB, where the observer additionally intercepts the action of the other agent and can choose the same action or another to be played in each round.
The \textbf{Upper Confidence Bound} (UCB) family of MAB algorithms balance exploration and exploitation by maintaining a statistic about the rewards per available action and calculate an exploration bonus per action. 
The \textbf{Risk-Averse Biased Upper Confidence Bound} (RAB UCB) MAB algorithm models human behavior by incorporating the risk-averse bias described by the \textit{CPT}. This model transforms observed rewards and probabilities, which results in a biased mean shift that changes the choice behavior of the agent: Concerning rewards, losses are weighted more strongly than gains. Additionally, differences between two large values are perceived as smaller than the same difference between two small values. Concerning probabilities, small probabilities are overweighted and large probabilities are underweighted.

\section{Algorithm}

The proposed methods works as follows. Without a training phase, after MAB $\mathcal{M}$ is instantiated with a given horizon $T$, the human policy $H$ is instantiated with a set of parameters for the CPT transformations $\alpha$, $\beta$, $\lambda$, $\gamma$, $\delta$ \cite{tversky1992advances}, and the rationality coefficient $\theta$ for the final noisy rational choice. The robot policy $R$ is instantiated with a model of the human, i.e., it has an assumption about the human behavior and an initial value for the estimated rewards $r_0$. $H$ has a descriptive statistic of previous rewards and their probabilities per arm, on which the CPT transformation is performed, and an arm pull counter for the exploration bonus. $R$ keeps a history of human choices, robot choices, and reward class per time step $\mathcal{H} = [(a^H_t, a^R_t, c_t), \dots]$.
Choosing an arm at time step $t$ is described in Alg. \ref{alg:hrteam}.
The main idea is that for each previous time step the probability distributions $\mathbf{P}_t$ of the reward classes per arm are known as well as the actual human choice $a_t^H$. Next, the probability distributions are transformed via CPT and in an optimization step (using an implementation of Powell's conjugate direction method \cite{powell1964efficient}), reward values are fitted to the reward classes to explain the human action each time step.
Finally, the inverse reward transformation is performed on these biased estimated reward values. The risk-averse bias is thus eliminated from the estimated reward values, which allows a more rational decision in the current time step.

\begin{algorithm}
\caption{Robot policy in H $\circ$ R Team}\label{alg:hrteam}
\begin{algorithmic}[1]
\Procedure{Choose Arm}{t}
\State create probability statistic $\boldsymbol{P}$ for $i \in [1, \dots, t-1]$ from  $\mathcal{H}$
\State perform probability bias on all $\boldsymbol{P}_i$, initialize $\boldsymbol{\hat{R}}$ with $r_0$
\State minimize $\sum_{i = 1}^{t-1} \text{argmax}_{a \in \mathcal{A}} \boldsymbol{P}_i \boldsymbol{\hat{R}} \neq a^H_i$
\State inverse reward transformation on $\boldsymbol{\hat{R}}$
\State choose $a^R_t = \text{argmax} \boldsymbol{P}_{t-1}\boldsymbol{\hat{R}}$
\State observe human choice $a^H_t$
\State H and R observe reward $r_t$ after $a^R_t$, update $\mathcal{H}$
\EndProcedure
\end{algorithmic}
\end{algorithm}

\section{Experiments}

The experiments seek to answer the following research questions: \textit{R1 risky-better}: Can the H-R team improve the performance of a risk-averse biased human policy in scenarios, where the risky option has a higher expected return? \textit{R2 safe-better}: Will the H-R team's performance deteriorate below the performance of the rational UCB policy when the lower variance option yields the higher expected return?

The experiments use a horizon $T =300$ MAB averaged over $N = 300$ trials and compare a baseline UCB policy, a RAB UCB human policy, and a human-robot team. The human policy is instantiated with $\theta = 1$, $\alpha = \beta = 0.5$, $\lambda = 2$, $\gamma = \delta = 0.5$ and the robot policy has knowledge of these parameters. For each trial, the policies are exposed to the same MAB by fixing the order that rewards per arm are sampled for all three actors. 
We fix two representative MABs with $N = 2$ arms and $M = 3$ different rewards for comparison: In \textit{risky-better}  MAB, a higher return can be achieved, when the arm with the more unsure and risky events is preferred. In the \textit{safe-better} MAB, the opposite is true.

Table \ref{exploratorytable} shows the returns of the two MAB settings for each agent. As expected, the H-R team moves the RAB UCB average return closer to the UCB agent. This means that in the risky-better MAB, the return improves, whereas the return deteriorates for the safe-better MAB. Both changes have approximately the same magnitude. Closing the gap to the UCB agent, who acts (and explores) rationally, is a positive property of the human-robot team. It is unclear if there is a way to keep the overly optimistic behavior when an actor finds itself in an environment where risk-aversion is indeed rewarded.

Performing a one-way ANOVA for the 3 agent types in the risky-better MAB reveals a significant difference:  F = 284.29, p = 0.0. A Tukey HSD post-hoc comparison shows that $\mu_{UCB}$ is significantly  higher than $\mu_{RABUCB}$, $p = 0.001$ and $\mu_{HRT}$, $p = 0.001$. However,  $\mu_{HRT} > \mu_{RABUCB}, p = 0.058$, was not significantly higher, therefore our assumption that the H-R team improves the performance (\textit{R1 risky-better}) is not fully supported by the data. 
Performing a one-way ANOVA for the three agent types in the safe-better MAB revealed a significant difference $F=18.73$, $p=0.0$. A Tukey HSD post-hoc comparison shows that $\mu_{RABUCB}$ outperformed $\mu_{UCB}$, $p=0.033$ and $\mu_{HRT}, p=0.012$. However, there was no difference between $\mu_{HRT}$ and $\mu_{UCB}$, $p=0.90$, which supports research question \textit{R2 safe-better}.
\begin{table}[]
\caption{Results for the comparison of two MABs ($N=300$, $T=300$) for three different agents.}
\begin{tabular}{llll}
\toprule
\textbf{MAB} & \textbf{Agent}              & \textbf{avg. return}                        &   \textbf{std.}  \\     \midrule
risky-better    & UCB           & 100.44    & 11.89  \\ 
                & RAB UCB  &   81.98  &  8.80 \\ 
                & HR Team        &  84.12   & 10.13 \\ \midrule
safe-better     & UCB          & 155.94    & 10.68  \\ 
                & RAB UCB & 161.41  & 10.89  \\ 
                & HR Team           & 159.18  & 11.35 \\ \bottomrule
\end{tabular}
\label{exploratorytable}
\end{table}

\section{Conclusion}
In this paper we motivated the \textit{Assistive Multi-Armed Bandit with Observable Reward Classes}. This formulation allows a direct observation of the variance of each arm in a MAB.
The problem is transformed from a preference learning problem of arm choices to a preference learning of different discrete rewards. This setting is a stepping stone to explore more complex human policies. Future work will include testing algorithms with a supervised training phase before the test phase in the expanded setting and reducing the error induced by the altered exploration behavior in the human policy.
\bibliographystyle{ACM-Reference-Format}
\bibliography{sample-base}
\end{document}